\pdfoutput=1

\documentclass[11pt]{article}

\usepackage[preprint]{acl}

\usepackage{times}
\usepackage{latexsym}
\usepackage{amsfonts, amsmath, amssymb, amsthm}

\usepackage[T1]{fontenc}

\usepackage[utf8]{inputenc}

\usepackage{microtype}

\usepackage{inconsolata}

\usepackage{graphicx}
\usepackage{float}

\usepackage{multirow}

\theoremstyle{definition}
\newtheorem{definition}{Definition}[section]

\title{LSR-Adapt: Ultra-Efficient Parameter Tuning with Matrix Low Separation Rank Kernel Adaptation}

\author{Xin Li \\
  Rutgers, the State University of \\New Jersey, NJ, USA \\
  \texttt{xl598@rutgers.edu} \\\And
  Anand D. Sarwate \\
  Rutgers, the State University of \\New Jersey, NJ, USA\\
  \texttt{anand.sarwate@rutgers.edu} \\}

\begin{document}
\maketitle
\begin{abstract}
Imposing an effective structural assumption on neural network weight matrices has been the major paradigm for designing Parameter-Efficient Fine-Tuning (PEFT) systems for adapting modern large pre-trained models to various downstream tasks. However, low rank based adaptation has become increasingly challenging due to the sheer scale of modern large language models. In this paper, we propose an effective kernelization to further reduce the number of parameters required for adaptation tasks. Specifically, from the classical idea in numerical analysis regarding matrix Low-Separation-Rank (LSR) representations, we develop a kernel using this representation for the low rank adapter matrices of the linear layers from large networks, named the Low Separation Rank Adaptation (LSR-Adapt) kernel. With the ultra-efficient kernel representation of the low rank adapter matrices, we manage to achieve state-of-the-art performance with even higher accuracy with almost half the number of parameters as compared to conventional low rank based methods. This structural assumption also opens the door to further GPU-side optimizations due to the highly parallelizable nature of Kronecker computations.
\end{abstract}

\section{Introduction}



Effectively designing structural assumptions is the key to the parameter-efficient approximation of network weight matrices. Low-Rank Adaptation (LoRA) \cite{lora} has been the pioneering method in PEFT that assumes a low-rank structure of the network weight matrices. However, despite its earlier successes, with an increasing number of parameters of modern day large language models, such simple structural assumption simply cannot effectively reduce the number of parameters into a manageable size with decent accuracy. To address this issue, various other structural assumptions have been proposed over the years \cite{qlora, dora, krona, kadaptation, peft}. Most of these methods, however, despite the effectiveness, lack solid theoretical reasoning of their structure design choices, and thus do not offer fine-grained control on the model performance. In this work, we provide a PEFT kernel based on the separable representation of matrices derived from the ideas in high dimensional numerical analysis to further decompose the factor matrices in various PEFT methods, coined as the \textit{Low Separation Rank Adaptation (LSR-Adapt)} kernel, which not only yields higher fine-tuning accuracy with even less trainable parameters, but also provides a solid theoretical foundation of this structural assumption for more control over the fine-tuning process.

In summary, the major contributions of this paper are as follows.
\begin{enumerate}
    \item Developing a structural assumption for PEFT based on a separable representation of matrices, which can be used as a kernel to further decompose the factor matrices of various PEFT methods, such as LoRA family methods \cite{lora, qlora}.
    \item Providing a theoretical analysis of the structure choice to give more insight for fine-tuning performance control.
    \item Experimental evaluations of our method as compared to other state-of-the-art PEFT methods against GLUE and SuperGLUE benchmarks \cite{glue, superglue}.
    \item Discussions on how this kernel structured computation can be parallelized using GPU, which can be interesting for further research in high-performance computing \cite{fast-kron}.
\end{enumerate}

\section{Related Works}

Numerous attempts have been done regarding Parameter-Efficient Fine-Tuning to adapt modern large language models to various applications. LoRA \cite{lora} has been one of the first major attempts in imposing efficient structural assumption on the neural network weight matrices of large models, subsequent research based on LoRA involves utilizing lower-precision quantization to harness the advantages of efficient calculations on lower-precision numbers offered by contemporary tensor core-based GPUs \cite{qlora}, and other form of weight decompositions with better semantic understanding of the weight matrices \cite{dora}. Further more, Kronecker product based factorizations of the weight matrices have also been studied to further reduce the parameter counts \cite{krona, kadaptation}, and \citeauthor{kadaptation} provides a mixture of low-rank and Kronecker factorization to achieve parameter-efficient tuning for vision models.

\section{Preliminaries}

To develop an efficient kernel to supercharge parameter-efficient tuning using separated representation of factor matrices, we first recall the generic definition of the separated representation in high-dimensional numerical analysis \cite{matrix-separated-representation},

\begin{definition}[The Separated Representation]
    Given an equation in $r$ dimensions ($r$ independent variables), we can try to approximate its solution $f$ by the following separation of variables,
    \begin{align}
        f(&x_1, \cdots, x_r)\nonumber\\
        &= \sum_{k = 1}^s \lambda_k \cdot g_k^{(1)}(x_1) \cdots g_k^{(r)}(x_r) + \mathcal{O}(\epsilon)
    \end{align}
    which is called a \textit{separated representation}, where $\mathcal{O}(\epsilon)$ is the desired asymptotic error proportional to $\epsilon$, $\{g_k^{(i)}(x_i)\}$ is the factor function for the $r$-dimensional variable $\boldsymbol{x} = \{x_1, \cdots, x_r\}$ at each dimension $i \in \{1, \cdots, r\}$ and $\lambda_k$ is a scaling factor for the $k$-th summation term where $k \in \{ 1, 2, \cdots, s \}$, and $s$ is called the \textit{separation rank}.
\end{definition}

This formulation effectively allows one to approximate a high-dimensional function $f$ with a linear complexity of $\mathcal{O}(r)$. Using this idea, we define the separated representation of matrices, by thinking an matrix $\boldsymbol{M} \in \mathbb{R}^{m_1 \times m_2}$ of dimension / rank $r \leq m_2$ (may not be full rank) as a discrete representation of an $r$-dimensional linear operator $\mathcal{M}$ on a rectangular domain of indices $(i, j) \in \mathbb{R}^{m_1 \times m_2}$, \textit{i.e.}, the matrix entries $M_{i, j} = \mathcal{M}(i, j)$, we can effectively extend the separated representation for $r$-dimensional functions to matrices.

\begin{definition}[The Matrix Separated Representation]
    For a given approximation error $\epsilon$, we can represent the matrix $\boldsymbol{M} \in \mathbb{R}^{m_1 \times m_2}$ as,
    \begin{equation}
        \boldsymbol{M} = \sum_{k = 1}^s \lambda_k \boldsymbol{M}_k^{(1)} \otimes \cdots \otimes \boldsymbol{M}_k^{(r)} + \mathcal{O}(\epsilon)
    \end{equation}
    with scalars $\lambda_1 \geq \cdots \geq \lambda_r > 0$, the integer $s$ the matrix \textit{separation rank}, and the factor matrix $\boldsymbol{M}_k^{(i)}$ is of dimension $m_{k, 1}^{(i)} \times m_{k, 2}^{(i)}$ and $\prod_{i = 1}^r m_{k, 1}^{(i)} = m_1,\: \prod_{i = 1}^r m_{k, 2}^{(i)} = m_2$ for all $k = 1, 2, \cdots, s$. In practice, we would like this separation rank term to be low for a parameter-efficient approximation, which leads to the matrix \textit{Low Separation Rank (LSR)} structure.
\end{definition}

The operator ``$\otimes$'' in the definition above denotes the Kronecker product. Specifically, for two matrices $\boldsymbol{U} \in \mathbb{R}^{u_1 \times u_2}, \boldsymbol{V} \in \mathbb{R}^{v_1 \times v_2}$, their Kronecker product, denoted as $\boldsymbol{U} \otimes \boldsymbol{V} \in \mathbb{R}^{(u_1 v_1) \times (u_2 v_2)}$, takes the format,
\begin{equation*}
    \boldsymbol{U} \otimes \boldsymbol{V} = \begin{bmatrix}
        U_{1, 1}\boldsymbol{U} &U_{1, 2}\boldsymbol{V} &\cdots &U_{1, u_2}\boldsymbol{V}\\
        U_{2, 1}\boldsymbol{U} &U_{2, 2}\boldsymbol{V} &\cdots &U_{2, u_2}\boldsymbol{V}\\
        \vdots  &\vdots &\ddots &\vdots\\
        U_{u_1, 1}\boldsymbol{U} &U_{u_1, 2}\boldsymbol{V} &\cdots &U_{u_1, u_2}\boldsymbol{V}
    \end{bmatrix}  \label{eq:mat-sep}
\end{equation*}

To gain a fine-grained control for the accuracy of the approximation, \citeauthor{matrix-separated-representation} proposed to use a condition number for the separated representation.

\begin{definition}[Condition Number of A Separated Representation]
    The condition number of \eqref{eq:mat-sep} is the ratio
    \begin{equation}
        \gamma = \frac{\left( \sum_{k = 1}^s \lambda_k^2 \right)^{1/2}}{\| \boldsymbol{M} \|_F}
    \end{equation}
    where $\| \cdot \|_F$ denotes the Frobenius norm.
\end{definition}

In a numerical computing system, we do not want $\gamma$ to be too large, a good rule of thumb to set the condition number would be to make it satisfy \cite{matrix-separated-representation},
\begin{equation}
    \gamma \mu \| \boldsymbol{M} \|_F \leq \epsilon
\end{equation}
where $\mu$ is the machine round-off, for instance, in a $16$-bit precision machine, the round-off number is $\mu = 2^{-11} \approx 4.88 \times 10^{-4}$. With this condition number, we can gain a fine-grained control over desired accuracy and dimensions of the factor matrices constrained by the numerical precision used during the finetuning process.




\begin{figure*}[hbt!]
    \centering
    \includegraphics[width=0.8\textwidth, height=17em]{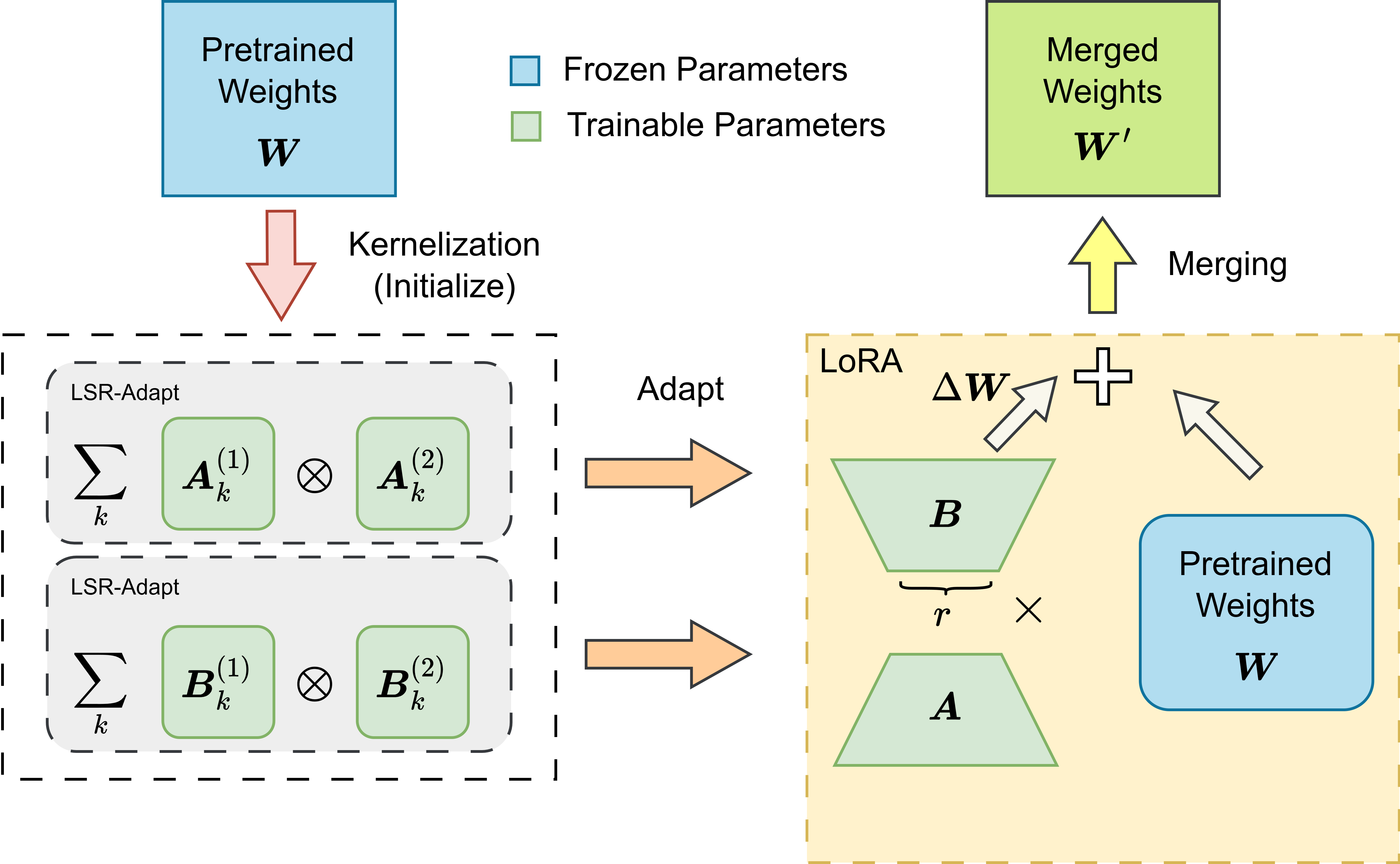}
    \caption{Overview of the working mechanism of LSR-Adapt kernel.}
    \label{fig:lsr-adapt}
\end{figure*}

\section{Our Approach}

\begin{table*}[hbt!]
  \centering
  \begin{tabular}{l|lll|lllll|l}
    \hline
    \multirow{2}{*}{\textbf{Method}} & \multicolumn{3}{c|}{\textbf{GLUE}} & \multicolumn{5}{c|}{\textbf{SuperGLUE}} & \multirow{2}{*}{\textbf{Average}} \\
    \cline{2-9}
    & \textbf{MRPC} & \textbf{SST-2} & \textbf{CoLA} & \textbf{RTE} & \textbf{CB} & \textbf{COPA} & \textbf{WSC} & \textbf{BoolQ} & \\
    \hline
    LoRA                & 74.51 & \textbf{94.43} & 83.32 & 68.23 & 76.79 & 57.39 & 63.46 & 75.41 & 74.19 \\
    KronA               & 76.57 & 94.12 & 82.59 & 67.92 & 79.23 & 56.48 & 63.46 & 74.97 & 74.42 \\
    KAdaptation         & 77.68 & 93.81 & 83.16 & 68.73 & 78.63 & 57.94 & 63.46 & 75.22 & 74.69 \\
    LSR-Adapt           & \textbf{80.88} & 94.27 & \textbf{83.41} & \textbf{68.95} & \textbf{82.14} & \textbf{60.32} & \textbf{63.46} & \textbf{75.72} & \textbf{76.14} \\
    \hline
  \end{tabular}
  \caption{\label{tab:glue+super-glue}
    Performance comparison of different adaptation methods on GLUE and SuperGLUE benchmark tasks.
    The best results for each task are bolded.
  }
\end{table*}

To develop an even more parameter efficient tuning mechanism, we are looking at a more parameter-efficient representation of the weight update matrix $\Delta \boldsymbol{W} \in \mathbb{R}^{w_1 \times w_2}$ for the weight matrix $\boldsymbol{W}  \in \mathbb{R}^{w_1 \times w_2}$ of the target network layer $\ell$,
\begin{equation}
    \boldsymbol{W} = \boldsymbol{W} + \Delta \boldsymbol{W},
\end{equation}
while a naive approach to adopt the matrix separable representation is to simply do,
\begin{equation}
    \Delta \boldsymbol{W} \approx \sum_{k = 1}^s \lambda_k \boldsymbol{W}_k^{(1)} \otimes \cdots \otimes \boldsymbol{W}_k^{(r)}
\end{equation}
where $r = \mathrm{rank}(\Delta \boldsymbol{W})$, while we can follow the common hypothesis in LoRA \cite{lora} where the weight update matrix is approximately low rank and use a rather small $r$, chaining even $r > 3$ Kronecker products can still be computationally expensive. Hence, we choose to take the low rank adapter matrices from LoRA,
\begin{equation}
    \Delta \boldsymbol{W} \approx \boldsymbol{A}\boldsymbol{B}
\end{equation}
for
\begin{equation}
    \boldsymbol{W}' = \boldsymbol{W} + \alpha \Delta \boldsymbol{W},
\end{equation}
where $\boldsymbol{A} \in \mathbb{R}^{w_1 \times r}$ and $\boldsymbol{B} \in \mathbb{R}^{r \times w_2}$ are the factor matrices, $\alpha$ is a scalar controlling the impact of the weight update matrix during adaptation, and structure the factor matrices $\boldsymbol{A}$, $\boldsymbol{B}$ with their matrix low-separation rank representations, \textit{i.e.}, the LSR-Adapt Kernel. Note that since $r$ is already a small value from the LoRA assumption, we can simply use two Kronecker factor matrices at each summation term of the LSR-Adapt kernel to achieve a decent reduction of parameter counts as compared to original LoRA,
\begin{align}
    \boldsymbol{A} &\approx \sum_{k = 1}^{s_A} \lambda_k^A \boldsymbol{A}_k^{(1)} \otimes \boldsymbol{A}_k^{(2)} \\
    \boldsymbol{B} &\approx \sum_{k = 1}^{s_B} \lambda_k^B \boldsymbol{B}_k^{(1)} \otimes \boldsymbol{B}_k^{(2)},
\end{align}
where $s_A$ and $s_B$ are respective separation ranks for factor matrices $\boldsymbol{A}$ and $\boldsymbol{B}$, which for simplified evaluation we set $s_A = s_B = s$, $\lambda_k^A$ and $\lambda_k^B$ are the corresponding scalar factors at summation term $k = 1, 2, \cdots, s$, which we will drop in the actual implementation and merge them into the $\alpha$ factor of the final low-rank factorization, the small Kronecker factor matrices take the shape $\boldsymbol{A}_k^{(i)} \in \mathbb{R}^{a_{k, 1}^{(i)} \times a_{k, 2}^{(i)}}$ for $i = \{1, 2\}$ and $\boldsymbol{B}_k^{(j)} \in \mathbb{R}^{b_{k, 1}^{(j)} \times b_{k, 2}^{(j)}}$ for $j = \{1, 2\}$, where,
\begin{align}
    a_{k, 1}^{(1)} \times a_{k, 1}^{(2)} &= w_1, &&a_{k, 2}^{(1)} \times a_{k, 2}^{(2)} = r \nonumber\\
    b_{k, 1}^{(1)} \times b_{k, 1}^{(2)} &= r, &&b_{k, 2}^{(1)} \times b_{k, 2}^{(2)} = w_2,
\end{align}
in practice we simply set $a_{k, 2}^{(1)} = b_{k, 1}^{(1)} = r^{(1)}$ and $a_{k, 2}^{(2)} = b_{k, 1}^{(2)} = r^{(2)}$ such that $r^{(1)} \times r^{(2)} = r$. Thus the weight update matrix takes the format,
\begin{align}
    \Delta &\boldsymbol{W} \approx \nonumber\\
    &\left( \sum_{k = 1}^{s} \boldsymbol{A}_k^{(1)} \otimes \boldsymbol{A}_k^{(2)} \right) \times \left( \sum_{k = 1}^{s} \boldsymbol{B}_k^{(1)} \otimes \boldsymbol{B}_k^{(2)} \right)
\end{align}
for
\begin{align}
    \boldsymbol{W}' &= \boldsymbol{W} + \alpha \Delta \boldsymbol{W} \nonumber\\
        &\approx \boldsymbol{W} + \alpha \left( \sum_{k = 1}^{s} \boldsymbol{A}_k^{(1)} \otimes \boldsymbol{A}_k^{(2)} \right) \times \nonumber\\
        &\qquad\qquad  \left( \sum_{k = 1}^{s} \boldsymbol{B}_k^{(1)} \otimes \boldsymbol{B}_k^{(2)} \right).
\end{align}
A simple diagram of this adaptation mechanism is shown in Figure \ref{fig:lsr-adapt}. Note that this is much more parameter-efficient as compared to the original low-rank factorization. Take a $768 \times 768$ network weight matrix for instance, if we set $r = 8$, we are looking at $2 \times 768 \times 8 = 12,288$ parameters, with an even higher rank $r = 16$, and assume balanced dimensions for the small kernel weight matrices, say, $\boldsymbol{A}_k^{(1)} \in \mathbb{R}^{32 \times 4}, \boldsymbol{A}_k^{(2)} \in \mathbb{R}^{24 \times 4}$, $\boldsymbol{B}_k^{(1)} \in \mathbb{R}^{32 \times 4}, \boldsymbol{B}_k^{(2)} \in \mathbb{R}^{24 \times 4}$, and separation rank $s = 16$, we can achieve a much lower parameter count $2 \times (32 \times 4 + 24 \times 4) \times 16 = 5,632$ while still maintaining a higher accuracy as we have found in fine-tuning experiments.

One can also show that this Kronecker-product based structure is amenable to parallel computation on modern power GPUs \cite{matrix-computations, fast-kron}. This enables potential development of custom CUDA kernels to further improve the training runtime, which is left to the future work.




\section{Experiments}

For our experiments \footnote{For detailed experimental setups and implementations, please feel free to check out our GitHub Repository: \url{https://anonymous.4open.science/r/lsr-adapt-7707}.}, we test our kernel for PEFT against both GLUE \cite{glue} and SuperGLUE benchmarks \cite{superglue} with RoBERTa model \cite{roberta}, the results are summarized in Table \ref{tab:glue+super-glue}. We train our model along with other baseline models using Hugging Face's \texttt{Trainer} framework with the default learning rate scheduler provided by the Transformers library, which is a linear scheduler with warmup \cite{lr-scheduler}. The model is optimized with a batch size of 256 for training and 64 for evaluation. For GLUE benchmark experiments, we train all the models for $20$ epochs and for the more challenging SuperGLUE benchmark experiments, we train all the models for $50$ epochs to get a more faithful comparison. Regarding the model hyperparameter set up, we set the LoRA rank as $8$ for all of our fine-tuning experiments, which leads to a parameter count of $2 \times 768 \times 8 = 12,288$ for the attention layer of dimension $768 \times 768$ with $\alpha = 32$. For our LSR-Kernel experiments, we set $r = 4$ and $\boldsymbol{A}_k^{(1)} \in \mathbb{R}^{32 \times 2}, \boldsymbol{A}_k^{(2)} \in \mathbb{R}^{24 \times 2}$, $\boldsymbol{B}_k^{(1)} \in \mathbb{R}^{32 \times 2}, \boldsymbol{B}_k^{(2)} \in \mathbb{R}^{24 \times 2}$, and separation rank $s = 16$, we can achieve a much lower parameter count $2 \times (32 \times 2 + 24 \times 2) \times 16 = 3,584$. All the other baseline methods follow the optimal settings given in the original papers \cite{krona, kadaptation}. From the results in Table \ref{tab:glue+super-glue}, we can see that our method still maintains a high performance for the PEFT benchmark tasks with almost $25\%$ of the LoRA parameters.

\section{Conclusion and Future Works}

In this paper, we have demonstrated the effectiveness of adopting the separable representations in PEFT tasks. Specifically, we have shown that by restructuring the LoRA factor matrices using matrix low separation rank representations, we can not only drastically reduce the number of trainable parameters, but also provide more robust fine-tuning accuracy. However, in this study, we did not fully utilize the favorable computational attributes of Kronecker products. This could enhance the efficiency of computation during training, and we plan to explore this in future research.

\newpage

\section{Limitations}

As discussed in the main paper, this work does not exploit the amenable computation properties of Kronecker products on modern tensor-core based GPUs, which might lead to further memory efficiency and faster training runtime, and potential robustness to low-precision training.

\medskip

\bibliography{main}

\appendix

\section{Appendix}
\label{sec:appendix}

\subsection{Basic Properties of Kronecker Products}
\label{subsec:appendix-1}

In this section we review some basic properties of the Kronecker products, which can be helpful in the case that we treat some matrix $\boldsymbol{A}$ as a block matrix whose the entries are all scalar multiplies of the same matrix \cite{matrix-computations}. Denote $\boldsymbol{A} \in \mathbb{R}^{m \times n}$ where $m = m_1 m_2,\: n = n_1 n_2$, then the matrix $\boldsymbol{A}$ is a \textit{Kronecker product} means that there exist two Kronecker factor matrices $\boldsymbol{B} \in \mathbb{R}^{m_1 \times n_1}$ and $\boldsymbol{C} \in \mathbb{R}^{m_2 \times n_2}$ such that,
\begin{equation}
    \boldsymbol{A} = \boldsymbol{B} \otimes \boldsymbol{C}.
\end{equation}

Some of the important Kronecker product properties include,
\begin{align}
    &\text{Transpose: } (\boldsymbol{B} \otimes \boldsymbol{C})^\top = \boldsymbol{B}^\top \otimes \boldsymbol{C}^\top \nonumber\\
    &\text{Product: } (\boldsymbol{B} \otimes \boldsymbol{C}) (\boldsymbol{D} \otimes \boldsymbol{E}) = \boldsymbol{B} \boldsymbol{D} \otimes \boldsymbol{C} \boldsymbol{E} \nonumber\\
    &\text{Associativity: } \boldsymbol{B} \otimes (\boldsymbol{C} \otimes \boldsymbol{D}) = (\boldsymbol{B} \otimes \boldsymbol{C}) \otimes \boldsymbol{D}. \nonumber
\end{align}

As for multiple Kronecker products, say $\boldsymbol{A} = \boldsymbol{B} \otimes \boldsymbol{C} \otimes \boldsymbol{D}$, one can regard it as a block matrix whose entries are block matrices. Specifically, for $(i, j)$-th block of $\boldsymbol{A}$, the value $B_{i, j} C_{k, l} \boldsymbol{D}$ is its $(k, l)$-th block.

\subsection{Understanding the Matrix Low Separation Rank Representation}
\label{subsec:appendix-2}

Here we provide a mathematical analysis on why a matrix low-separation rank representation is an effective approximation mechanism for matrices. Suppose we have a matrix $\boldsymbol{A} \in \mathbb{R}^{m \times n}$ with $\mathrm{rank}(\boldsymbol{A}) = r$, we would like to show that it admits the following approximation,
\begin{equation}
    \boldsymbol{A} = \sum_{k = 1}^s \lambda_k \boldsymbol{A}^{(1)}_k \otimes \cdots \otimes \boldsymbol{A}^{(r)}_k + \mathcal{O}(\epsilon),
\end{equation}
with a separation rank $s$ and $\boldsymbol{A}_k^{(i)} \in \mathbb{R}^{m_i \times n_i}$, where $\prod_{i = 1}^r m_i = m,\: \prod_{i = 1}^r n_i = n$. We start from the fact that $\mathrm{rank}(\boldsymbol{A}) = r$, and thus there exist vectors $\boldsymbol{u}_1, \cdots, \boldsymbol{u}_r \in \mathbb{R}^m$ and $\boldsymbol{v}_1, \cdots, \boldsymbol{v}_r \in \mathbb{R}^n$ such that,
\begin{equation}
    \boldsymbol{A} = \sum_{k = 1}^r \boldsymbol{u}_k \boldsymbol{v}_k^\top.
\end{equation}
Then for each rank-$1$ matrix $\boldsymbol{u}_k \boldsymbol{v}_k^\top \in \mathbb{R}^{m \times n}$, we can do the reshaping,
\begin{align}
    \boldsymbol{u}_k &= \boldsymbol{u}_k^{(1)} \otimes \cdots \otimes \boldsymbol{u}_k^{(r)},\\
    \boldsymbol{v}_k &= \boldsymbol{v}_k^{(1)} \otimes \cdots \otimes \boldsymbol{v}_k^{(r)},\\
\end{align}
where each vector $\boldsymbol{u}_k^{(i)} \in \mathbb{R}^{m_i}, \boldsymbol{v}_k^{(i)} \in \mathbb{R}^{n_i}$. Thus from the basic Kronecker properties mentioned in \ref{subsec:appendix-1} we have,
\begin{align}
    \boldsymbol{u}_k \boldsymbol{v}_k^\top &= \left( \otimes_{i = 1}^r \boldsymbol{u}_k^{(i)} \right) \left( \otimes_{i = 1}^r \boldsymbol{v}_k^{(i)} \right)^\top \nonumber\\
    &= \left( \otimes_{i = 1}^r \boldsymbol{u}_k^{(i)} \right) \left( \otimes_{i = 1}^r \left( \boldsymbol{v}_k^{(i)} \right)^\top \right) \nonumber\\
    &= \bigotimes_{i = 1}^r \left( \boldsymbol{u}_k^{(i)} \left( \boldsymbol{v}_k^{(i)} \right)^\top \right).
\end{align}
To see why the last equality in the above derivations works, consider the simpler example where we wish to compute
\begin{equation}
    \left( \boldsymbol{u}^{(1)} \otimes \boldsymbol{u}^{(2)} \otimes \boldsymbol{u}^{(3)} \right) \left( \boldsymbol{v}^{(1)} \otimes \boldsymbol{v}^{(2)} \otimes \boldsymbol{v}^{(3)} \right), \nonumber
\end{equation}
if we define the substitutions $\boldsymbol{U} = \boldsymbol{u}^{(1)} \otimes \boldsymbol{u}^{(2)}$ and $\boldsymbol{V} = \boldsymbol{v}^{(1)} \otimes \boldsymbol{v}^{(2)}$, with the Kronecker properties in \ref{subsec:appendix-1} we have the above equation becomes,
\begin{align}
    \left( \boldsymbol{U}  \otimes \boldsymbol{u}^{(3)} \right) \left( \boldsymbol{V}  \otimes \boldsymbol{v}^{(3)} \right) = \left( \boldsymbol{U}\boldsymbol{V} \right) \otimes \left( \boldsymbol{u}^{(3)} \boldsymbol{v}^{(3)} \right) \nonumber
\end{align}
where
\begin{align}
    \boldsymbol{U} \boldsymbol{V} &= \left( \boldsymbol{u}^{(1)} \otimes \boldsymbol{u}^{(2)} \right) \left( \boldsymbol{v}^{(1)} \otimes \boldsymbol{v}^{(2)} \right) \nonumber\\
    &= \left( \boldsymbol{u}^{(1)} \boldsymbol{v}^{(1)}\right) \otimes \left( \boldsymbol{u}^{(2)} \boldsymbol{v}^{(2)}\right).
\end{align}
Then we substitute this back, yielding
\begin{align}
    &\left( \boldsymbol{u}^{(1)} \otimes \boldsymbol{u}^{(2)} \otimes \boldsymbol{u}^{(3)} \right) \left( \boldsymbol{v}^{(1)} \otimes \boldsymbol{v}^{(2)} \otimes \boldsymbol{v}^{(3)} \right)\nonumber\\
    &\quad = \left( \boldsymbol{u}^{(1)} \boldsymbol{v}^{(1)}\right) \otimes \left( \boldsymbol{u}^{(2)} \boldsymbol{v}^{(2)}\right) \otimes \left( \boldsymbol{u}^{(3)} \boldsymbol{v}^{(3)} \right).
\end{align}
Or in simplified notations,
\begin{equation}
    \left( \otimes_{i = 1}^3 \boldsymbol{u}^{(i)} \right) \left( \otimes_{i = 1}^3 \boldsymbol{v}^{(i)} \right) = \bigotimes_{i = 1}^3 \left( \boldsymbol{u}^{(i)} \boldsymbol{v}^{(i)} \right).
\end{equation}

Then if we define,
\begin{equation}
    \boldsymbol{A}_k^{(i)} \triangleq \boldsymbol{u}_k^{(i)} \left( \boldsymbol{v}_k^{(i)} \right)^\top \in \mathbb{R}^{m_i \times n_i},
\end{equation}
we essentially have
\begin{equation}
    \boldsymbol{u}_k \boldsymbol{v}_k^\top = \boldsymbol{A}_k^{(1)} \otimes \cdots \otimes \boldsymbol{A}_k^{(r)}.
\end{equation}
To make the computations more controllable, we can set all the factor matrices $\boldsymbol{A}_k^{(i)}$ to be of unit norm (\textit{e.g.}, Frobenius norm or operator norm) and factor out a scalar factor $\lambda_k$,
\begin{equation}
    \boldsymbol{u}_k \boldsymbol{v}_k^\top = \lambda_k \boldsymbol{A}_k^{(1)} \otimes \cdots \otimes \boldsymbol{A}_k^{(r)}.
\end{equation}
Then we have the form
\begin{align}
    \boldsymbol{A} &= \boldsymbol{u}_k \boldsymbol{v}_k^\top \nonumber\\
    &= \sum_{k = 1}^r \lambda_k \boldsymbol{A}_k^{(1)} \otimes \cdots \otimes \boldsymbol{A}_k^{(r)},
\end{align}
however, in practice, $\boldsymbol{A}$ might not be low rank and attaining the actual $r$ can be expensive, hence if we instead set the approximate rank $r \leq \mathrm{rank}(\boldsymbol{A})$, we have the following approximation
\begin{equation}
    \boldsymbol{A} = \sum_{k = 1}^s \lambda_k \boldsymbol{A}^{(1)}_k \otimes \cdots \otimes \boldsymbol{A}^{(r)}_k + \mathcal{O}(\epsilon),
\end{equation}
where integer $s \geq r$ is the separation rank and $\epsilon$ is the approximation error.
\end{document}